\documentclass[conference]{IEEEtran}
\IEEEoverridecommandlockouts

\usepackage{booktabs}
\usepackage{amssymb}
\usepackage{amsmath} 
\usepackage{graphicx}
\usepackage{acronym}
\usepackage{siunitx}
\usepackage{multirow}
\usepackage[hidelinks]{hyperref}
\usepackage{makecell}
\usepackage{multirow}
\usepackage{siunitx}
\usepackage{subcaption}
\usepackage{acronym}%

\def\BibTeX{{\rm B\kern-.05em{\sc i\kern-.025em b}\kern-.08em
    T\kern-.1667em\lower.7ex\hbox{E}\kern-.125emX}}

\graphicspath{ {./figures/} }

\begin{document}

\title{Magnetic Indoor Localization through CNN~Regression and Rotation Invariance
}

\author{
\IEEEauthorblockN{Helge Rosé, Konstantin Klipp, Tom Koubek}
\IEEEauthorblockA{\textit{Fraunhofer Institute for Open}\\
\textit{Communication Systems (FOKUS)} \\
Berlin, Germany \\
\{helge.rose, konstantin.klipp, \\tom.koubek\}@fokus.fraunhofer.de
}
\and
\IEEEauthorblockN{Bernd Schäufele, Ilja Radusch}
\IEEEauthorblockA{\textit{Daimler Center for Automotive} \\
\textit{Information Technology Innovations (DCAITI)}\\
\textit{Technical University Berlin}\\
Berlin, Germany\\
\{bernd.schaeufele, ilja.radusch\}@dcaiti.com
}
}

\makeatletter

\makeatother

\maketitle

\begin{abstract}
Indoor positioning is an essential technology for a wide range of applications in GNSS-denied environments, including indoor navigation and IoT systems. Combining convolutional neural networks (CNNs) and magnetic field-based features offers a low-cost, infrastructure-free solution for precise positioning. While magnetic fingerprints are a promising approach for indoor positioning, models trained on raw 3D magnetometer data are highly sensitive to device orientation. We address this by using two rotation invariant features derived from the 3D magnetic field: the norm ($M_n$) and the projection onto the gravity axis ($M_g$). We train a lightweight 7-layer dilated CNN (MagNetS/XL) on magnetic sequences to directly regress $(x,y)$ positions. Using the MagPie dataset (three buildings, handheld trajectories), we systematically evaluate fixed and random rotations of test and/or train data. 
Raw 3D inputs $(M_x,M_y,M_z)$ exhibit isotropic error increases under fixed \SI{90}{\degree} rotations and further degrade with growing random rotations. In contrast, 2D $(M_n,M_g)$ inputs maintain rotation invariant accuracy and surpass the 3D inputs once rotation exceeds building-specific thresholds for three reference buildings: \SI{0}{\degree} for Loomis (large), \SI{5}{\degree} for Talbot (medium), and \SI{6}{\degree} for CSL (small).
MagNetXL achieves or exceeds state-of-the-art accuracy on the MagPie dataset, and MagNetS delivers similar performance with roughly one third of the parameters, favoring mobile deployment. These results show that the robustness gained from rotation invariant inputs outweighs the loss of input dimensionality in realistic usage, allowing mapping and localization without orientation alignment or added infrastructure.
\end{abstract}


\acrodef{BLE}{Bluetooth Low Energy}
\acrodef{CNN}{Convolutional Neural Network}
\acrodef{GNSS}{Global Navigation Satellite Systems}
\acrodef{GPS}{Global Positioning System}
\acrodef{IoT}{Internet of Things}
\acrodef{LSTM}{Long Short-Term Memory}
\acrodef{MAE}{Mean Absolute Error}
\acrodef{RNN}{Recurrent Neural Network}
\acrodef{SVM}{Support Vector Machine}

\section{Introduction}

The emergence of \ac{GNSS}, such as the \ac{GPS}, has facilitated widespread and widely accessible positioning solutions in outdoor environments. Navigation systems for vehicles and pedestrians are ubiquitous and readily accessible through various smartphone applications.

In contrast, navigation applications are not universally available in indoor environments. A significant challenge lies in accurate localization within buildings, as \ac{GNSS} typically requires a direct line-of-sight between the receiver device and the satellites. Since \ac{GNSS} signals cannot penetrate walls, their availability indoors is limited. However, in the context of high-precision localization, ensuring a seamless transition between outdoor and indoor environments can be very beneficial, particularly for blind individuals \cite{wortmann2024}. In addition, indoor localization also enables other location-based services like the tracking of \ac{IoT} assets or patients in hospitals \cite{maloney2007tool}.

To address this limitation, multiple technologies have been explored for indoor localization. Most existing systems rely on technologies such as radio, optical, acoustic, or magnetic information, or a combination of them \cite{schyga2022meaningful}. Most of these technologies face challenges related to infrastructure costs and maintenance efforts. For instance, radio-based systems such as \ac{BLE} and WiFi require the installation of multiple access points or beacons. Besides installation costs, constant maintenance is necessary to keep the system operational. Other approaches exploiting optical features, which have lower infrastructure requirements, are prone to degradation under different illumination levels and require line of sight.

Thus, in our previous research, we examined the magnetic characteristics of buildings \cite{klipp2025}. Magnetic-based solutions rely solely on the Earth’s magnetic field, which is essentially available everywhere.  The vast majority of buildings contain large metal parts like steel beams or radiators. These structures affect the magnetic field forming a specific magnetic pattern while moving through the object.

To this end, in \cite{klipp2025} we employed different neural networks and demonstrated that they can estimate a highly accurate position based on three-dimensional magnetic patterns. For better comparison, we used a common public dataset for this problem called MagPie \cite{hanley2022magnetic}. However, we have determined that the dataset is not realistic in its raw format for all our focused applications. This limitation arises from the assumption of a more or less constant phone orientation depending on the walking direction. In real-world scenarios, such as indoor navigation, these tasks are predominantly executed on smartphones or other handheld devices. Given that users typically hold their devices while walking or have them in their pocket, a constant orientation of the magnetometer cannot be expected. Users are likely to rotate the device about its yaw, pitch, and roll axes, thereby affecting the vector direction of the magnetic field. The same issue arises if \ac{IoT} devices are tracked using this approach, as the orientation of the tracker or the device it is attached to may vary while in motion. 

To enhance the robustness of the magnetic localization framework, this paper introduces a neural network-based localization system that only uses rotation invariant features of the magnetic vector. These features consist of the norm of the magnetic field strength and the projection of the magnetic field components in the sensor reference frame to the vertical axis of the Earth's gravity field. We also introduce two new \ac{CNN}-based localization models, MagNetXL and MagNetS. We use the models to show the influence of sensor rotations on the results with the raw sensor values in comparison to the rotation invariant features.

This paper is structured as follows. In \autoref{sec:related}, we review related work in this field. The concept of rotation invariant magnetic features is introduced in \autoref{sec:rotinv}. The preparation of the dataset and the architecture of our networks are explained in \autoref{sec:data}. In \autoref{sec:results}, the evaluation metrics and our results are presented. Finally, \autoref{sec:conclusion} provides a conclusion and outlines future research directions.

\section{Related Work}
\label{sec:related}

In addition to the infrastructure-based localization techniques using Bluetooth, WiFi, or Ultrawideband anchors in the building, there are infrastructure-free approaches as well. \cite{zhang2021indoor} and \cite{klipp2021multidimensional} 
use inertial sensors with existing map data to reduce drift characteristics by particle or Kalman filters. \cite{Kendall_2015_ICCV, wang2021continual} and \cite{dfnet} utilize camera images to detect natural features and structures in the scene and perform image retrieval or end-to-end pose regression to find the best match in a prerecorded image or feature map. However, these approaches require more or less robust features and line of sight.

Approaches using the Earth's magnetic patterns mostly rely on a prerecorded fingerprinting map to localize a measured pattern as well. For this purpose, a wide range of traditional algorithms are used, such as k-nearest neighbors \cite{KNN_Li2016}, \ac{SVM} \cite{SVM_Abdou2016,SVM_Chriki2017} and decision trees \cite{DecisionTrees_YIM2008} even if some of them actually focus on WiFi fingerprints \cite{ouyang2023}. In recent years, neural network-based solutions like \cite{ouyang2023,ashraf2020,abid2021} proposed \ac{CNN}s and \ac{RNN}s, especially \ac{LSTM} architectures for localization.
\cite{jang2017}, \cite{son2025universal}, \cite{yadav2020indoor}, and \cite{wang2021} all employed variations of \ac{RNN} respectively \ac{LSTM} models in their work. The complexity of these models ranges from basic \ac{RNN}s (\cite{jang2017}) to more sophisticated architectures like the hierarchical \ac{LSTM} proposed by \cite{wang2021}. The approaches in \cite{ouyang2023}, \cite{ashraf2020}, and \cite{abid2021} utilize \ac{CNN}-based architectures in their work. Recent research has also explored the combination of different neural network architectures to exploit their respective strengths. Both \cite{antsfeld2021} and \cite{fernandes2020} use more advanced hybrid approaches that integrate \ac{CNN}s with \ac{RNN}s or \ac{LSTM}s to build and process multichannel visual representations of magnetic data. These approaches combine the spatial processing capabilities of \ac{CNN}s with the sequential learning of \ac{RNN}s. However, these architectures sometimes come with the drawback that a spatial starting point is needed. Thus, the combination of \ac{CNN} and \ac{RNN}, e.g., in \cite{antsfeld2021}, cannot solve the lost robot problem.

In our latest work \cite{klipp2025}, we evaluated different \ac{CNN}- and \ac{LSTM}-based solutions for a classification-based and a regression-based location estimation. A Bayesian optimizer was used to tune the model on a subset of the data per building. Thus, for each building a specific model structure is generated. As with most state-of-the-art solutions, the magnetic input pattern consists of the three-dimensional magnetic vector over a certain time period gathered by a sliding window approach. For a comparable evaluation the public MagPie dataset \cite{hanley2022magnetic} was used, as it includes all necessary data at a suitable and realistic measuring frequency of \SI{50}{\hertz} in differently sized buildings. The best networks in \cite{klipp2025} achieve average mean errors between \SI{0.2}{\meter} and \SI{2}{\meter} depending on the building size, outperforming state of the art networks on the same dataset. 

Although the MagPie dataset was collected using two handheld devices, a smartphone and a tango tablet respectively for the ground truth, the data do not show much variance in rotation and only limited influence from phone movements. As each path typically has two walking directions, the training, validation, and test data only contain these two rotations. Thus, the work presented in this paper addresses the gap to a more realistic scenario regarding rotations.  

In an earlier work \cite{klipp2018rotation}, we implemented a grid-based probabilistic Bayesian filter to match invariant magnetic features against a map. The present study now employs a \ac{CNN} that incorporates rotation invariant magnetic features to derive an absolute position through regression analysis.

\section{Magnetic Indoor Localization and Rotation Invariance}
\label{sec:rotinv}

If indoor localization is to be based solely on data from the Earth's magnetic field, the orientation of the device has a major influence on the accuracy of the localization \cite{Haverinen2013}. The values of the magnetic field components depend on this orientation in relation to the direction of the Earth's magnetic field. The more local magnetic fields disturb the Earth's globally north directed magnetic pattern, the more accurate magnetic localization becomes \cite{Haverinen2013}. However, the more disturbance they cause, the more difficult it becomes to determine the absolute orientation of the device using the Earth's magnetic field.

In magnetic localization, the magnetic field is recorded twice: first, when the magnetic map of the building is created, and then during the actual localization. It is very likely that the recorded values of the map differ from the measurements taken during the localization phase, because the orientation of the sensor device changes between map recording and localization mode. Therefore, the direct use of magnetic field components causes significant challenges in mitigating orientation dependence \cite{klipp2018rotation, Haverinen2013}, leading to poor localization results.

One approach to address this challenge is to use rotation invariant quantities of the magnetic field. The magnetic field has three components $(M_x,M_y,M_z)$, and of course the number of rotation invariant quantities is smaller, namely two quantities (see \autoref{sec:invfeatures}).

The central aim of this work is to quantify how much the expected accuracy gain from input rotation invariance can compensate for the loss in accuracy caused by reducing the input dimensionality.

\subsection{Options for random rotation}
\label{sec:somerot}
To test the above-mentioned hypothesis, we rotated the input data in two different ways. First, we rotated the data by a constant angle, (see \autoref{ite:fir}). Then, we rotated the data in a varying manner (see \autoref{ite:sec} and \autoref{ite:thi}).
\begin{enumerate}
    \item\label{ite:fir} Rotate the test data by a fixed angle, representing a device in localization mode, with a permanently different device orientation than during recording, e.g., the user carries the device in their pocket.
    \item\label{ite:sec} Randomly rotate the test data only, representing a definite map recording by a robot or a specific trained person and a randomly directed localization mode. This localization mode roughly simulates e.g. a user holding the phone in the hand while varying its orientation during a walk.
    \item\label{ite:thi} Randomly rotate both the test and training data to simulate a randomly directed device at map recording and a randomly directed device in localization mode. This scenario is the most likely one if the map data are gathered during normal smartphone use in a crowd-based mapping approach.
\end{enumerate}

The randomly varying rotations are applied, following the normal distribution $N(0,\sigma)$. This simulates the natural behavior of a user who performs many small wobbles and a few larger turns. Note that the probability of $2\sigma$ movements is already very small. The time interval between two random rotations is $T$. In most cases, we choose $T=\SI{1}{\second}$. Linear interpolation is used between random rotations, i.e., at time steps $t=nT,n=0..K$ the random rotation angle $\phi(t) \in N(0,\sigma)$ is sampled, and the intermediate values of $\phi(\tau), \tau\in (t,t_1), t=nT, t_1=(n+1)T$ are linearly interpolated between $\phi(t)$ and $\phi(t_1)$. Additionally, fixed rotations are applied over the entire time period.

\section{Data preparation and neural networks}
\label{sec:data}

Here we follow the approach presented in \cite{klipp2025} and focus on data sequences, because the same magnetic field value can occur for different positions, and thus the localization based on a single magnetic data point is ambiguous. The sequence length remains as a parameter for evaluation, as its effect is of high relevance for the performance and for the possible applications as well. The tests in \cite{klipp2025} show that an overlapping sliding window of $W=200$ is suitable in most cases. In contrast to consecutive, non-overlapping ones, this avoids losing information between windows, as the target position always corresponds to the last data point in the sequence. In the mentioned work, we used the $3D$ raw magnetic field vector $M = (M_x,M_y,M_z)$ as input data. Both in \cite{klipp2025} and in this work we use all floors in the MagPie dataset from three buildings of different sizes (CSL - small, Talbot - medium, Loomis - large). To eliminate effects that are not related to this study, we only use the data recordings for handheld devices without obvious outliers. We removed a few trials from training and evaluation (namely trials 8 to 11 in the Loomis building) due to time-sync issues in the given MagPie dataset.

\subsection{Rotation invariant features}
\label{sec:invfeatures}

We are testing our hypothesis by using rotation invariant magnetic features instead of the raw magnetic components.
The magnetometers of the mobile devices used for the MagPie dataset measure the magnetic field strength $M(r)$ at the position $r = (lon,lat,alt)$ w.r.t. the world reference frame (longitude, latitude, altitude) or in UTM coordinates $r = (X,Y,Z)$. The field strength \(M(r) = (M_x, M_y, M_z) \in \mathbb{R}^3\) is expressed in the sensor’s coordinate frame, with components measured along the device’s local \(x\)-, \(y\)- and \(z\)-axes. To establish rotation invariance, it is necessary to consider the group of 3-dimensional rotations $O(3)$ given by all transformations of the vector space $\mathbb{R}^3$ leaving the inner product invariant. Thus, the norm of a vector is also invariant of rotations. This means that the first $O(3)$ rotation invariant feature of the magnetic field strength is its norm $M_n := |M|$.

The subgroup $O(2)$ of 2-dimensional rotations about a local vertical axis of the world reference frame leaves the vertical component of the magnetic field strength invariant. The local vertical axis of the UTM world reference frame is parallel to the unit vector $g$ of the Earth's gravity field detectable by the accelerometer of the sensor device. The projection $M_g := M\cdot g$ of the magnetic field components in the sensor reference frame to the vertical axis $g$ of the Earth's gravity field is the second rotation invariant feature. Accordingly, we use $(M_n, M_g)$ as the rotation invariant 2D input data of the neuronal networks.

\subsection{Networks by hyperparameter search}
\label{secsec:cnn}
In \cite{klipp2025} the neural networks are kept as flexible as possible, and the best model architecture and optimal training parameters were identified by conducting a hyperparameter search using Bayesian optimization on an undersampled training set. The basic model structures are \ac{CNN} and \ac{LSTM} architectures. Both types are applied to a classification and regression problem with the $3D$ raw magnetic field vector $M = (M_x,M_y,M_z)$ as non-rotation invariant input data. The regression directly predicts the position vector $r=(x,y,z=0)$. The regression approach generally outperforms the classification method in terms of positioning accuracy and precision. The regression approach offers a more straightforward implementation, as it does not require dividing the building into distinct grid cells. Thus, in this paper we consider the regression approach only.

In \cite{klipp2025}, the hyperparameter search of \ac{LSTM} networks performed better than the \ac{CNN} approach, resulting in different optimal network architectures for each building and configuration. Since our goal is to investigate the rotational invariance of the networks, it is advantageous to consider only a fixed network architecture for the various input data rotation approaches mentioned in \autoref{sec:somerot}. Therefore, we manually optimized the 7-layer \ac{CNN} architecture MagNet, which outperforms both \cite{klipp2025} and most reference networks considered therein in terms of the \ac{MAE}.

We consider both network types, from \cite{klipp2025} and MagNet, and also use the \ac{MAE} between the ground truth and the predicted positional coordinates to assess and compare the performance of the networks. The \ac{MAE} is given by

\begin{equation}
\label{eq:MAE}
\text{MAE} = \frac{1}{n} \sum_{i=1}^{n} \left| y_i - \hat{y}_i \right|
\end{equation}

where $y_i$ are the ground truth coordinates and $\hat{y}_i$ are the predicted coordinates of the networks.

\begin{table}[b!]
\centering
\caption{Comparison of magnetic localization approaches from~\cite{antsfeld2021} and~\cite{klipp2025} to MagNetXL on original MagPie set without rotation. }
\begin{tabular}{@{}lcccc@{}}
\toprule
\textbf{Building} &
\makecell{ \textbf{\cite{antsfeld2021} (CNN+RNN)} \\ \textbf{\ac{MAE} (m)} } & \makecell{ \textbf{\cite{klipp2025} } \\ \textbf{\ac{MAE} (m)}} & \makecell{\textbf{MagNetS} \\ \textbf{ \ac{MAE} (m)}} &
\makecell{\textbf{MagNetXL} \\ \textbf{ \ac{MAE} (m)}} \\ \midrule
CSL                           & 0.30                 & 0.22  & 0.22  &  \textbf{0.21} \\
Talbot                        & 1.06                 & 0.82  & 0.66  &  \textbf{0.64}\\ 
Loomis                        & \textbf{1.07}        & 1.98  & 2.13  &  1.96 \\
\bottomrule
\end{tabular}
\label{tab:research_comparison}
\end{table}

\subsection{MagNet - a 7-layer CNN for magnetic indoor localization}
\label{secsec:magnet}

To optimize the \ac{CNN} from \cite{klipp2025}, we designed a customized 7-layer \ac{CNN}. The design is based on the following consideration:

A suitable Conv1D design for a time series of $W=200$ should use small kernels and increasing dilation rates so that early layers learn local patterns while deeper layers achieve a receptive field covering the full sequence. For $n$~layers, kernel size $k$ and exponentially increasing dilation rates (i.e.,~$1, 2, 4, 8, 16, $ etc.), the receptive field length can be approximated by $R \approx 1+(k-1)(2^n - 1)$, thus to get a length of $R>200$, in our approach we have to select $n=7$, $k=3$ yielding $R\approx255$. Filters typically increase with depth to capture progressively more complex features. Using a global pooling layer at the end efficiently summarizes all learned features.

Thus, the number of features increases from 32 to 128 (MagNetS) or 256 (MagNetXL). The \ac{CNN} layers are implemented with dilation ranging from 1 to 64 in order to achieve a receptive field corresponding to the input sequence length of 200. In test runs we found that kernel sizes $k>3$ result in higher precision; thus, we increase $k=5 ... 20$ in each successive layer to mitigate gridding artifacts, especially when small kernels are combined with high dilation rates. To implement nonlinear regression to the output $(x,y)$, the \ac{CNN} layers are followed by a nonlinear fully connected layer of size 64 and a linear fully connected layer of size 2. The smaller network -- MagNetS -- has 360k parameters, while the larger one -- MagNetXL -- has approximately 1M.

Both networks achieve comparable accuracy, with MagNetXL performing slightly better in difficult situations. For normal cases, and especially on mobile devices, MagNetS, which is three times smaller, is preferable. \autoref{tab:research_comparison} shows that both MagNet networks outperform \cite{klipp2025} and most reference networks considered in \cite{antsfeld2021}. In contrast to the CNN+RNN version from \cite{antsfeld2021}, the networks in \cite{klipp2025}, MagNetS and MagNetXL, also solve the lost robot problem as no starting position is fed into the network.

\section{Results}
\label{sec:results}
In the following section, we show the advantage of rotation invariant input data. 
In \autoref{secsec:axis}, we apply a fixed rotation on the three different axes of the $3D$ test data separately, to analyze the dependence of the axis of rotation on the \ac{MAE}.
In \autoref{secsec:rancnn} and \autoref{secsec:ranmag}, we compare the \ac{MAE} dependence on fixed and varying rotations of $2D$ and $3D$ input data.

\subsection{The dependence of the Mean Absolute Error on the axis of rotation}
\label{secsec:axis}

The results of fixed rotations around all axis to the $3D$ test data are shown in \autoref{tab:csl200-fix88} for MagNetS. This simulates a user carrying their cell phone upright in their pocket. The rotation of the test data by a fixed angle represents a device in localization mode, with a permanently different device direction than during recording. Rotation by almost \SI{90}{\degree} around all three axes individually and also together shows an approximate error of \SI{14}{\meter} in all cases. There is no directional dependence, i.e., the error is significant and isotropic.

\begin{table}[b!]
    \centering
    \caption{Dependence of the Mean Absolute Error (in \si{\meter}) on the axis of rotation. Building: CSL, Network: MagNetS with $3D$ input, test data rotated by \SI{88}{\degree} around the x-, y-, z-axis and all three axes simultaneously. In any case, the error is approximately \SI{14}{\meter}, i.e., there is no directional dependence — the error is isotropic.}
    \begin{tabular}{@{}ccccc@{}}
        \toprule
         \textbf{x \ac{MAE} (m)} & \textbf{y \ac{MAE} (m)} & \textbf{z \ac{MAE} (m)} & \textbf{xyz \ac{MAE} (m)} \\ \midrule
               13.60         & 13.44  & 14.38 &  13.45\\
        \bottomrule
    \end{tabular}
    \label{tab:csl200-fix88}
\end{table}

\begin{figure*}
    \centering
    \begin{subfigure}[b]{0.45\linewidth}  
        \centering
        \includegraphics[height=0.77\linewidth]{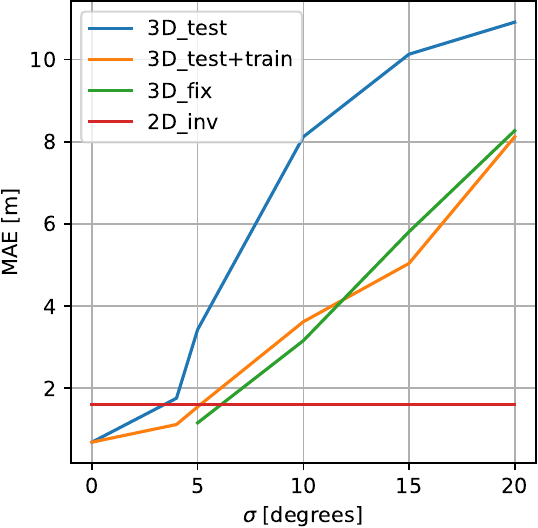}
        \caption{The dependence of the Mean Absolute Error on random rotations with the CNN by \cite{klipp2025}. Building: CSL, Network: CNN by \cite{klipp2025}. 3D\_test: 3D input data $(M_x,M_y,M_z)$, only test data rotated by N(0,$\sigma$) distributed angles around all axes. 3D\_test+train: test and train data rotated by N(0,$\sigma$). 3D\_fix: test and train data rotated by a fixed angle $\sigma$. 2D\_inv: 2D input data $(M_g,M_n)$, which are invariant with respect to rotations.}
        \label{fig:csl200-cnn}    
    \end{subfigure}
    \hspace{1cm}
    \begin{subfigure}[b]{0.45\linewidth}    
            \centering
        \includegraphics[height=0.77\linewidth]{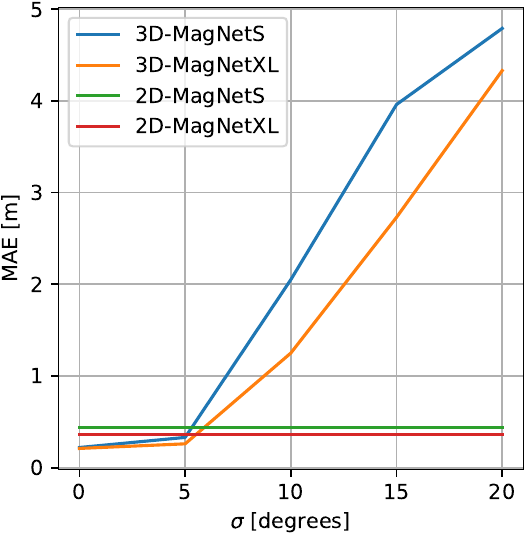}
        \caption{The dependence of the Mean Absolute Error on random rotations with MagNetS/XL. Building: CSL, Network: MagNetS and MagNetXL. 3D\_MagNetS/XL: 3D input data $(M_x,M_y,M_z)$, test and train data rotated by N(0,$\sigma$) distributed angles around all axes. 2D\_MagNetS/XL: 2D input data $(M_g,M_n)$, which are invariant with respect to rotations.\vspace{0.35cm}}%
        \label{fig:csl200-cnn7}    
    \end{subfigure}
    \caption{Comparison of the dependence of the Mean Absolute Error on random rotations with the CNN by \cite{klipp2025} and  MagNetS/XL.}
    \label{fig:csl200}
\end{figure*}

\begin{figure*}
    \centering
    \begin{subfigure}[b]{.45\linewidth}    
            \centering
        \includegraphics[height=0.77\linewidth]{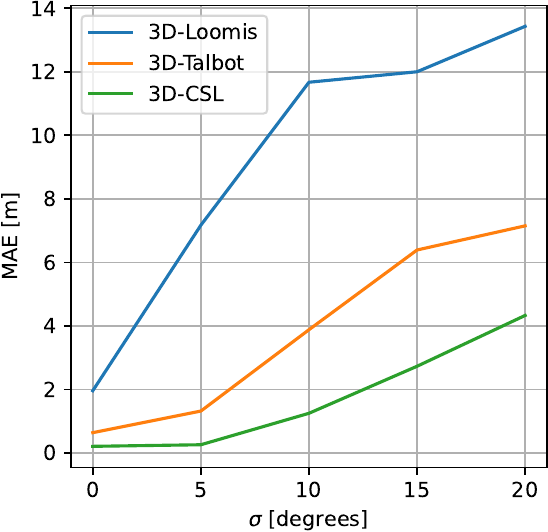}
        \caption{3D input data $(M_x,M_y,M_z)$, test and train data rotated by N(0,$\sigma$) distributed angles around all axes.}
        \label{fig:cnn7a}
    \end{subfigure}
    \hspace{1cm}
    \begin{subfigure}[b]{.45\linewidth}
            \centering
        \includegraphics[height=0.77\linewidth]{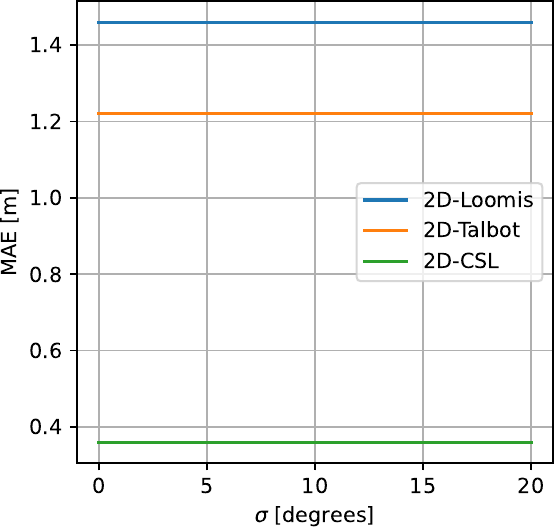}
         \caption{2D input data $(M_g,M_n)$, which are invariant with respect to rotations.}
        \label{fig:cnn7b}
    \end{subfigure}
    \caption{The dependence of the Mean Absolute Error on building size. Building: Loomis-big, Talbot-medium, CSL-small, Network: MagNetXL. Note the different scales for MAE with 3D input data in \autoref{fig:cnn7a} and 2D input data in \autoref{fig:cnn7b}.}
    \label{fig:cnn7}
\end{figure*}

\subsection{The dependence of the Mean Absolute Error on random rotations for the CNN presented in \cite{klipp2025}}
\label{secsec:rancnn}

The results of fixed (Option \ref{ite:fir}) and random (Options \ref{ite:sec} and \ref{ite:thi}) rotations of the 3D input data $(M_x,M_y,M_z)$ for the reference \ac{CNN} architecture by \cite{klipp2025} in the smallest building CSL are shown in Figure \ref{fig:csl200-cnn}. The graph {\em 3D\_test} shows the result for the 3D input data $(M_x,M_y,M_z)$, with only the test data rotated by N(0,$\sigma$) distributed angles around all axes (Option \ref{ite:sec}). Option \ref{ite:thi} where test and train data are rotated by N(0,$\sigma$) is shown by {\em 3D\_test+train}. The random rotations are drawn from normal distribution $N(0,\sigma)$. This simulates the natural behavior of a user who performs many small wobbles and a few larger turns. The time interval between two random rotations is $T=\SI{1}{\second}$. Linear interpolation is used between the random rotation times. In {\em 3D\_fix} the test and train data are rotated by a fixed angle $\sigma$ and applied over the entire time period (Option \ref{ite:fir}). The 2D input data $(M_g,M_n)$ are used in the rotation invariant case {\em 2D\_inv}. The input data and thus the results are invariant with respect to rotations.

It is clearly visible that the more the input data are rotated, the higher the \ac{MAE} gets, using the 3-dimensional rotation variant features. At the same time, of course the MAE of the rotation invariant features solution stays constant over all rotations $\sigma$.
Case {\em 3D\_test} shows the largest increase in error, up to 11m for random rotations with $\sigma=\SI{20}{\degree}$. From an angle of \SI{4}{\degree}, we can see that the advantage of rotation invariant 2D input data with $MAE=\SI{1.61}{\meter}$ surpasses the non-invariant 3D data with $MAE=\SI{1.76}{\meter}$.\\%

If both the training and test data are rotated (case {\em 3D\_test+train}), the error increases more gradually up to approximately \SI{8}{\meter}. The threshold angle at which the rotation invariant 2D data performs better than the 3D input data is \SI{5}{\degree}, which is similar to the {\em 3D\_test} case. This means that random disturbance of the data during data acquisition and application only results in a slightly slower increase in error than a disturbance of the application alone. The threshold angle at which the disturbance becomes effective remains approximately the same in both cases from \SI{5}{\degree}.

The case of a fixed rotation ({\em 3D\_fix}) has a linear dependency comparable to that in case {\em 3D\_test+train}. This shows that the randomness of the disturbance is not decisive, but only the magnitude of the rotation angle influences the error. Also in this case, the threshold value is approximately \SI{5}{\degree}.

\subsection{The dependence of the Mean Absolute Error on random rotations for MagNet}
\label{secsec:ranmag}

\autoref{fig:csl200-cnn7} shows the results of random rotations of the 3D input data $(M_x,M_y,M_z)$ and the invariant 2D input data $(M_g,M_n)$ in the smallest building CSL for MagNet. In all cases the test and training data are randomly rotated by a normal distribution $N(0,\sigma)$. As before, the time interval between two random rotations is $T=\SI{1}{\second}$ and a linear interpolation is used between the random rotation times. The rotations are applied for both networks MagNetS and MagNetXL. The cases {\em 3D-MagNetS/XL} show the dependency of the error for the non-invariant input data $(M_x,M_y,M_z)$ and {\em 2D-MagNetS/XL} for the rotation invariant 2D input data $(M_g,M_n)$.

In the non-invariant cases {\em 3D-MagNetS/XL}, the error increases more slowly than in the case of the \ac{CNN} in \cite{klipp2025}. Here, we have a maximal error of approximately \SI{4.5}{\meter} instead of \SI{8}{\meter}. The larger network MagNetXL outperforms the smaller one (MagNetS), but the maximal gain in error is only \SI{1.2}{\meter}. at an angle of \SI{15}{\degree}. For small rotation angles (up to \SI{5}{\degree}) both networks perform similarly well with an error of \SI{0.3}{\meter} at \SI{5}{\degree}.

For the rotation invariant 2D input data $(M_g,M_n)$, the larger MagNetXL network performs slightly better, with an error of \SI{0.36}{\meter}, than the smaller MagNetS, with \SI{0.44}{\meter}. Because of the invariance, the error is constant over the full range. The threshold angle, at which the rotation invariant 2D data performs better than the 3D input data, with \SI{5.5}{\degree}, is very similar to the case of the \ac{CNN} in \cite{klipp2025}.

\subsection{The Advantage of Rotation Invariance }

The results of the dependence of accuracy on rotational disturbances in the input data of the \ac{CNN} in \cite{klipp2025} and MagNet clearly show that, above a threshold rotation angle, the rotation invariant 2D input data $(M_g,M_n)$ enable better accuracy in magnetic localization. \autoref{fig:cnn7} and \autoref{tab:threshold} show the dependence of this threshold angle on the size of the building.

\begin{table}
\centering
\caption{Threshold value of the rotation angle for the advantage of rotation invariant 2D input data $(M_n, M_g)$ of MagNetXL for different building sizes. }
\begin{tabular}{@{}lcccc@{}}
\toprule
\textbf{Building} & \textbf{Threshold angle (\SI{}{\degree})} & \textbf{3D MAE (m) } & \textbf{2D MAE (m)} \\ \midrule
Loomis-big      & 0         & 1.96  & 1.46 \\
Talbot-medium   & 5         & 1.32  & 1.22 \\ 
CSL-small       & 6         & 0.36  & 0.36 \\
\bottomrule
\end{tabular}
\label{tab:threshold}
\end{table}

For the largest building (Loomis), the rotation invariant 2D input data $(M_g,M_n)$ achieve better accuracy than the 3D data $(M_x,M_y,M_z)$, even without rotation disturbances, i.e., the threshold value of the rotation angle is \SI{0}{\degree}. In addition, the invariant 2D data for both the large and medium-sized building (Talbot) achieve a comparatively small error of around \SI{1.3}{\meter}. For the small building (CSL), the error is around \SI{0.4}{\meter}, which is three times smaller.

The results show (see \autoref{tab:threshold}) that the threshold value of the rotation angle is inversely proportional to the size of the building. This indicates that in the case of smaller buildings, the networks are still able to easily learn small rotational disturbances and thus compensate for their influence below the threshold angle. This is possible up to an angle of \SI{6}{\degree} for the small building (CSL) and \SI{5}{\degree} for the medium-sized building (Talbot). In the case of the large building, the network is no longer able to learn all rotation disturbances in the data, and thus the rotation invariant 2D data $(M_g,M_n)$ always outperform the rotation-disturbed 3D data $(M_x,M_y,M_z)$.

\section{Conclusion}
\label{sec:conclusion}

This work examines the rotation invariance of magnetic input data for \ac{CNN}s in the task of magnetic indoor localization to answer the question: How effectively can the use of rotation invariant 2D input data compensate for rotation errors and thus achieve increased localization accuracy, and how large is the impact of the loss of dimensionality from the raw 3D data $(M_x,M_y,M_z)$ to the rotation invariant 2D data $(M_n, M_g)$?

We introduced two rotation invariant features derived from the 3D magnetometer signal -- the norm of the magnetic field ($M_n$) and its projection onto the gravity axis ($M_g$) -- and trained a lightweight 7-layer \ac{CNN} (MagNetS/XL) on sequences of length $W=200$. Across the MagPie dataset buildings, we showed that models using raw 3D inputs $(M_x,M_y,M_z)$ are highly sensitive to device orientation. Fixed rotations near \SI{90}{\degree} lead to isotropic error increases (e.g., $\sim$\SI{4}{\meter} MAE), and even small random rotations degrade accuracy. In contrast, using the features $(M_n, M_g)$ keeps the error constant across rotations and surpasses 3D inputs once the device rotation exceeds a building size-dependent threshold: \SI{0}{\degree} for Loomis (large), ~\SI{5}{\degree} for Talbot (medium), and ~\SI{6}{\degree} for CSL (small). Additionally, MagNetXL achieves or exceeds the state-of-the-art accuracy on MagPie, while MagNetS offers similar performance with substantially lower complexity, making it preferable for mobile deployment.

Practically, rotation invariant inputs remove the need to align device orientation between mapping and localization, improve robustness for handheld usage, and simplify system setup without extra infrastructure. The results indicate that the gain from rotational robustness outweighs the loss of dimensionality in most realistic scenarios, especially in larger environments where learning orientation disturbances becomes harder.

Our current solution faces certain limitations, including reliance on sufficiently distinctive magnetic fingerprints, potential errors in gravity estimation, device-specific magnetometer biases, and the focus on a single public dataset (handheld trajectories). Performance still depends on coverage and quality of mapping data and may vary with building construction and dynamic changes. However, as most of them affect the position certainty, they can be mitigated by sensor fusion, e.g., with accelerometers.

Future work will address trajectory-level inference and uncertainty estimation to achieve higher robustness and accuracy. For usage in an indoor navigation system \cite{wortmann2024}, also extension to floor detection and further on-device optimization for real-time application can be considered.

\section*{References}
\renewcommand{\section}[2]{}%
\bibliographystyle{ieeetr}
\bibliography{references}

@inproceedings{jang2017,
  title={{Geomagnetic field based indoor localization using recurrent neural networks}},
  author={Jang, Ho Jun and Shin, Jae Min and Choi, Lynn},
  booktitle={{GLOBECOM 2017-2017 IEEE Global Communications Conference}},
  pages={1--6},
  year={2017},
  organization={IEEE}
}

@article{son2025universal,
  title={{Universal Vector Calibration for Orientation-Invariant 3D Sensor Data}},
  author={Son, Wonjoon and Choi, Lynn},
  journal={Sensors},
  volume={25},
  number={15},
  pages={4609},
  year={2025},
  publisher={MDPI}
}

@article{ashraf2020,
  title={{MINLOC: Magnetic field patterns-based indoor localization using convolutional neural networks}},
  author={Ashraf, Imran and Kang, Mingyu and Hur, Soojung and Park, Yongwan},
  journal={IEEE Access},
  volume={8},
  pages={66213--66227},
  year={2020},
  publisher={IEEE}
}

@article{fernandes2020,
  title={{An infrastructure-free magnetic-based indoor positioning system with deep learning}},
  author={Fernandes, Let{\'\i}cia and Santos, Sara and Barandas, Mar{\'\i}lia and Folgado, Duarte and Leonardo, Ricardo and Santos, Ricardo and Carreiro, Andr{\'e} and Gamboa, Hugo},
  journal={Sensors},
  volume={20},
  number={22},
  pages={6664},
  year={2020},
  publisher={MDPI}
}

@inproceedings{abid2021,
  title={{Improved CNN-based magnetic indoor positioning system using attention mechanism}},
  author={Abid, Mahdi and Compagnon, Paul and Lefebvre, Gr{\'e}goire},
  booktitle={{2021 International Conference on Indoor Positioning and Indoor Navigation (IPIN)}},
  pages={1--8},
  year={2021},
  organization={IEEE}
}

@inproceedings{yadav2020indoor,
  title={{Indoor space classification using cascaded LSTM}},
  author={Yadav, Rohan Kumar and Bhattarai, Bimal and Jiao, Lei and Goodwin, Morten and Granmo, Ole-Christoffer},
  booktitle={2020 15th IEEE Conference on Industrial Electronics and Applications (ICIEA)},
  pages={1110--1114},
  year={2020},
  organization={IEEE}
}

@article{ouyang2023,
  title={{Magnetic-field-based indoor positioning using temporal convolutional networks}},
  author={Ouyang, Guanglie and Abed-Meraim, Karim and Ouyang, Zuokun},
  journal={Sensors},
  volume={23},
  number={3},
  pages={1514},
  year={2023},
  publisher={MDPI}
}

@ARTICLE{KNN_Li2016,
  author={Li, Dong and Zhang, Baoxian and Li, Cheng},
  journal={IEEE Internet of Things Journal}, 
  title={{A Feature-Scaling-Based $k$-Nearest Neighbor Algorithm for Indoor Positioning Systems}}, 
  year={2016},
  volume={3},
  number={4},
  pages={590-597},
  doi={10.1109/JIOT.2015.2495229}}

@INPROCEEDINGS{SVM_Abdou2016,
  author={Abdou, Ashraf Sayed and Aziem, Mostafa Abdel and Aboshosha, Ashraf},
  booktitle={{2016 Sixth International Conference on Digital Information Processing and Communications (ICDIPC)}}, 
  title={{An efficient indoor localization system based on Affinity Propagation and Support Vector Regression}}, 
  year={2016},
  volume={},
  number={},
  pages={1-7},
  doi={10.1109/ICDIPC.2016.7470782}}

@INPROCEEDINGS{SVM_Chriki2017,
  author={Chriki, Amira and Touati, Haifa and Snoussi, Hichem},
  booktitle={{2017 13th International Wireless Communications and Mobile Computing Conference (IWCMC)}}, 
  title={{SVM-based indoor localization in Wireless Sensor Networks}}, 
  year={2017},
  volume={},
  number={},
  pages={1144-1149},
  doi={10.1109/IWCMC.2017.7986446}}

@inproceedings{antsfeld2021,
  title={{Magnetic field sensing for pedestrian and robot indoor positioning}},
  author={Antsfeld, Leonid and Chidlovskii, Boris},
  booktitle={{2021 International Conference on Indoor Positioning and Indoor Navigation (IPIN)}},
  pages={1--8},
  year={2021},
  organization={IEEE}
}

@article{wang2021,
  title={{A hierarchical LSTM-based indoor geomagnetic localization algorithm}},
  author={Wang, Liying and Luo, Haiyong and Wang, Qu and Shao, Wenhua and Zhao, Fang},
  journal={IEEE Sensors Journal},
  volume={22},
  number={2},
  pages={1227--1237},
  year={2021},
  publisher={IEEE}
}

@phdthesis{hanley2022magnetic,
  title={{Magnetic-field-based navigation: Empirical evidence of robust methods over contemporary methods}},
  author={Hanley, David},
  year={2022},
  school={University of Illinois at Urbana-Champaign}
}

@inproceedings{wortmann2024,
  title={{Enhanced Accessibility for Mobile Indoor Navigation}},
  author={Johannes Wortmann and Bernd Schaeufele and Konstantin Klipp and Ilja Radusch and Katharina Blaß and Thomas Jung},
  booktitle={{2024 14th International Conference on Indoor Positioning and Indoor Navigation (IPIN)}},
  year={2024},
  organization={IEEE}
}

@inproceedings{klipp2018rotation,
  title={{Rotation-invariant magnetic features for inertial indoor-localization}},
  author={Klipp, Konstantin and Ros{\'e}, Helge and Willaredt, Jonas and Sawade, Oliver and Radusch, Ilja},
  booktitle={{2018 International Conference on Indoor Positioning and Indoor Navigation (IPIN)}},
  pages={1--10},
  year={2018},
  organization={IEEE}
}

@INPROCEEDINGS{klipp2025,
  author={Klipp, Konstantin and Blumenthal, Edgar and Eckardt, Marten and Windirsch, Jasper and Schäufele, Bernd and Wortmann, Johannes and Radusch, Ilja},
  booktitle={{2025 11th International Conference on Computing and Artificial Intelligence (ICCAI)}}, 
  title={{Comparing CNN and LSTM Networks for Magnetic Localization of IoT Devices and Pedestrian Tracking}}, 
  year={2025},
  volume={},
  number={},
  pages={549-555},
  doi={10.1109/ICCAI66501.2025.00088}
}

@article{DecisionTrees_YIM2008,
author = {Yim, Jaegeol},
title = {{Introducing a decision tree-based indoor positioning technique}},
year = {2008},
issue_date = {February, 2008},
publisher = {Pergamon Press, Inc.},
address = {USA},
volume = {34},
number = {2},
issn = {0957-4174},
url = {https://doi.org/10.1016/j.eswa.2006.12.028},
doi = {10.1016/j.eswa.2006.12.028},
journal = {Expert Syst. Appl.},
month = feb,
pages = {1296–1302},
numpages = {7}
}

@misc{Haverinen2013,
 title={{Utilizing magnetic field based navigation}},
 author={Haverinen, J.},
 year={US Patent US2013/0179074, 2013}
 }

@article{schyga2022meaningful,
  title={{Meaningful test and evaluation of indoor localization systems in semi-controlled environments}},
  author={Schyga, Jakob and Hinckeldeyn, Johannes and Kreutzfeldt, Jochen},
  journal={Sensors},
  volume={22},
  number={7},
  pages={2797},
  year={2022},
  publisher={MDPI}
}

@inproceedings{klipp2021multidimensional,
  title={{Multidimensional In-and outdoor pedestrian tracking using OpenStreetMap data}},
  author={Klipp, Konstantin and Kisand, Armin and Wortmann, Johannes and Radusch, Ilja},
  booktitle={2021 International Conference on Indoor Positioning and Indoor Navigation (IPIN)},
  pages={1--8},
  year={2021},
  organization={IEEE}
}

@article{zhang2021indoor,
  title={{An Indoor Localization Method Based on the Combination of Indoor Map Information and Inertial Navigation with Cascade Filter}},
  author={Zhang, Yushuai and Guo, Jianxin and Wang, Feng and Zhu, Rui and Wang, Liping},
  journal={Journal of Sensors},
  volume={2021},
  number={1},
  pages={7621393},
  year={2021},
  publisher={Wiley Online Library}
}

@InProceedings{Kendall_2015_ICCV,
author = {Kendall, Alex and Grimes, Matthew and Cipolla, Roberto},
title = {{PoseNet: A Convolutional Network for Real-Time 6-DOF Camera Relocalization}},
booktitle = {Proceedings of the IEEE International Conference on Computer Vision (ICCV)},
year = {2015}
}

@InProceedings{dfnet,
  title={{DFNet: Enhance Absolute Pose Regression with Direct Feature Matching Supplementary}},
  author={Chen, Shuai and Li, Xinghui and Wang, Zirui and Prisacariu, Victor A},
  booktitle={European Conference on Computer Vision},
  pages={1--17},
  year={2022},
  organization={Springer}
}

@inproceedings{wang2021continual,
  title={{Continual learning for image-based camera localization}},
  author={Wang, Shuzhe and Laskar, Zakaria and Melekhov, Iaroslav and Li, Xiaotian and Kannala, Juho},
  booktitle={Proceedings of the IEEE/CVF International Conference on Computer Vision},
  pages={3252--3262},
  year={2021}
}

@inproceedings{maloney2007tool,
  title={{A tool for improving patient discharge process and hospital communication practices: the Patient Tracker}},
  author={Maloney, Christopher G and Wolfe, Douglas and Gesteland, Per H and Hales, Joe W and Nkoy, Flory L},
  booktitle={AMIA Annual Symposium Proceedings},
  volume={2007},
  pages={493},
  year={2007}
}

\end{document}